\documentclass[letterpaper]{article}
\usepackage{aaai}
\usepackage{times}
\usepackage{helvet}
\usepackage{courier}

\usepackage{graphicx} 
\usepackage{float}

\usepackage{amsmath}
\usepackage{amssymb} 

\nocopyright 

\begin{document}
%
\title{Topology-Aware CLIP Few-shot Learning}
\author{Dazhi Huang\\
huangdesmond48@gmail.com\\
}
\maketitle
\begin{abstract}
\begin{quote}
Efficiently adapting large Vision-Language Models (VLMs) like CLIP for few-shot learning poses challenges in balancing pre-trained knowledge retention and task-specific adaptation. Existing methods often overlook valuable structural information within the VLM's latent space. We introduce a topology-aware tuning approach integrating Representation Topology Divergence (RTD) into the Task Residual (TR) framework. By explicitly aligning the topological structures of visual and text representations using a combined RTD and Cross-Entropy loss, while freezing base VLM encoders, our method enhances few-shot performance. We optimize only lightweight Task Residual parameters, effectively leveraging topological information. Across 6 diverse benchmark datasets, our approach demonstrates significant gains, achieving an average accuracy improvement of 1-2\% over relevant baseline methods in few-shot settings. This work presents an effective strategy to boost VLM few-shot capabilities by incorporating topological alignment.
\end{quote}
\end{abstract}

\section{Introduction}

Vision-Language Models (VLMs) have emerged as a powerful paradigm in artificial intelligence, bridging the gap between visual perception and natural language understanding. These models are typically trained on vast datasets comprising billions of image-text pairs scraped from the web, enabling them to learn rich and transferable semantic representations \cite{radford_learning_2021,li_blip-2_2023,alayrac_flamingo_2022,pmlr-v162-li22n}. By learning to associate visual concepts with their textual descriptions at scale, prominent VLMs such as CLIP \cite{radford_learning_2021}, BLIP \cite{pmlr-v162-li22n}, BLIP-2 \cite{li_blip-2_2023}, and Flamingo \cite{alayrac_flamingo_2022} have demonstrated remarkable abilities in capturing nuanced semantic information directly from raw multimodal data. This large-scale pre-training imbues the models with a strong foundation of general-purpose knowledge, effectively aligning visual and linguistic modalities within a shared embedding space.

Leveraging these powerful pre-trained representations for downstream tasks, particularly in few-shot scenarios where labeled data is scarce, has motivated the development of various efficient transfer learning (ETL) techniques. Mainstream approaches strive to adapt VLMs while preserving the valuable prior knowledge learned during pre-training. For instance, prompt tuning methods like CoOp \cite{zhou2022conditional} introduce learnable context vectors into the input prompts, effectively creating task-specific prompts. Adapter-style tuning approaches, such as CLIP-Adapter \cite{Gao2023CLIPAdapter}, insert lightweight modules, often with residual connections, into the VLM architecture to adapt features. Tip-Adapter \cite{Zhang2022TipAdapter} offers a training-free alternative by constructing a key-value cache model from the few-shot examples to modulate predictions. Methods like TaskRes \cite{yu2023task} take a different route by explicitly preserving the original VLM classifier and introducing tunable, prior-independent parameters as a task-specific residual. The central challenge in few-shot VLM adaptation lies in effectively utilizing the rich prior knowledge encoded within the pre-trained model while accurately inferring the posterior knowledge distribution specific to the target task from only a handful of examples.


The inherent structure of the shared image-text latent space, however, remains largely unexplored by current ETL methods, which often focus on instance-level or pairwise relationships. Topological Data Analysis (TDA), particularly through the lens of persistent homology, offers a novel perspective by providing tools to compare the multi-scale topological structures of data manifolds. Metrics derived from TDA, such as cross-barcodes \cite{NEURIPS2021_3bc31a43} and Representation Topology Divergence (RTD) \cite{pmlr-v162-barannikov22a}, have proven effective in quantifying structural differences between deep learning representations. Previous works have demonstrated the benefits of incorporating topological constraints; for example, Topological Autoencoders \cite{pmlr-v119-moor20a} and RTD Autoencoders \cite{trofimov2023learning} showed improved representation learning by aligning the topology of input and latent spaces. Similarly, TopoKD \cite{kim2024do} leveraged persistence images to indirectly match topological structures between teacher and student models for knowledge distillation. More directly relevant to VLMs, recent work explicitly constrained the structural equivalence of image and text latent manifolds using 0-dimensional persistent homology, achieving more effective ETL by aligning homology persistence between modalities \cite{zhang2024homology}. We hypothesize that enforcing topological consistency between image and text representations can significantly enhance VLM few-shot learning. In data-scarce settings, the topology of the latent space, shaped by contrastive pre-training like CLIP's which explicitly aligns modalities, encodes rich posterior knowledge. Specifically, 0-dimensional persistent homology captures semantic clustering information, including both intra-cluster similarity and inter-cluster separation. Furthermore, 1-dimensional homology can encode cyclic or periodic structures arising from continuous variations within related concepts, such as different breeds of dogs or cats within the same superclass. Capturing and aligning these multi-scale topological features offers a promising direction for improving the generalization and accuracy of VLMs when adapted with limited data.

Motivated by the potential of TDA to capture rich structural information within latent spaces, this paper focuses on leveraging Representation Topology Divergence (RTD) to enhance the few-shot learning performance of VLMs. We employ a differentiable RTD-based loss function, specifically adapted to align the topological structures of image and text representations within the VLM's shared embedding space. By explicitly enforcing topological consistency between modalities during the efficient transfer learning process, we aim to improve the model's ability to generalize from limited data. The main contributions of this work are summarized as follows:
\begin{itemize}
    \item We are the first to introduce higher-dimensional topological features, quantified by RTD, as a constraint for few-shot learning in vision-language models, moving beyond simpler topological measures.
    \item We employ a differentiable loss function derived from RTD, specifically tailored to align the multi-scale topological structures of cross-modal representations (image and text) within VLMs during efficient adaptation.
    \item Through extensive experiments on multiple few-shot classification benchmarks, we demonstrate that our topology-guided alignment method significantly boosts performance compared to existing state-of-the-art efficient transfer learning techniques.
\end{itemize}

\section{Background}

\subsection{Persistent Homology}

Persistent Homology (PH) is a central technique within Topological Data Analysis (TDA) designed to quantify and analyze the multi-scale topological structure inherent in data. Given a point cloud $P$ sampled from some underlying space, PH employs the concept of filtration to understand how its topology changes across different proximity scales.

A common way to construct a filtration from a point cloud $P$ residing in a metric space is by using the Vietoris-Rips (VR) complex. For a given scale parameter $\epsilon \ge 0$, the VR complex, denoted $\mathcal{R}_{\epsilon}(P)$, is a simplicial complex whose vertices are the points in $P$. A set of vertices $\{p_0, p_1, \dots, p_k\} \subseteq P$ forms a $k$-simplex in $\mathcal{R}_{\epsilon}(P)$ if the distance between any pair of vertices $p_i, p_j$ in the set is less than or equal to $\epsilon$. As $\epsilon$ increases, more simplices are included, resulting in a sequence of nested complexes: $\mathcal{R}_{\epsilon_0} \subseteq \mathcal{R}_{\epsilon_1} \subseteq \dots \subseteq \mathcal{R}_{\epsilon_n}$, where $0 = \epsilon_0 < \epsilon_1 < \dots < \epsilon_n$. This sequence is known as the VR filtration.

\begin{figure}[ht]
    \centering
    \includegraphics[width=0.85\linewidth]{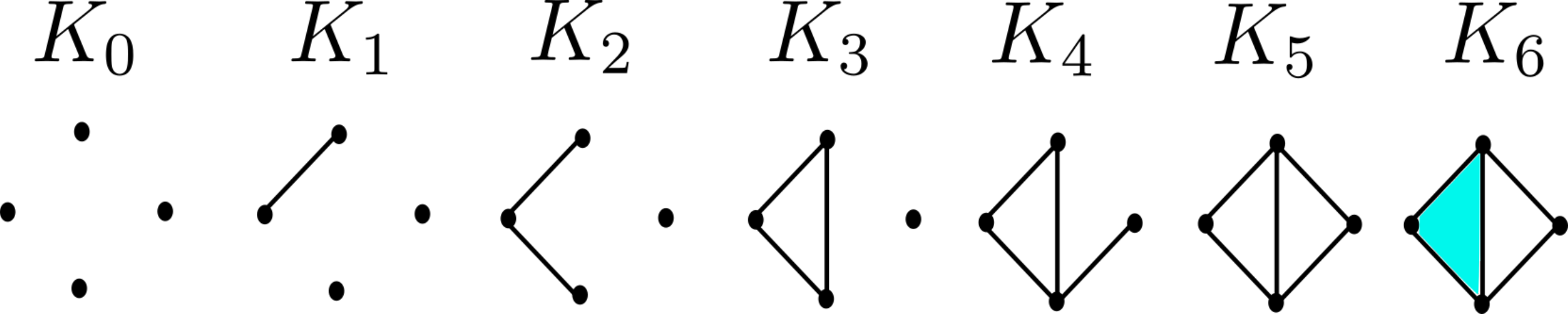}
    \caption{A filtration of a complex from a point cloud of four points.}
    \label{fig:complex1}
\end{figure}

\begin{center}
\begin{tabular}{|c|c|c|c|c|c|c|c|}
  \hline
  $H_p$ & $K_0$ & $K_1$ & $K_2$ & $K_3$ & $K_4$ & $K_5$ & $K_6$   \\ \hline
  0 & $\mathbb{Z}^4$ & $\mathbb{Z}^3$ & $\mathbb{Z}^2$ & $\mathbb{Z}^2$ & $\mathbb{Z}$ & $\mathbb{Z}$ & $\mathbb{Z}$  \\
  1 & 0 & 0 & 0 & $\mathbb{Z}$ & $\mathbb{Z}$ & $\mathbb{Z}^2$  & $\mathbb{Z}$   \\ \hline
  Betti(p) & $K_0$ & $K_1$ & $K_2$ & $K_3$ & $K_4$ & $K_5$ & $K_6$   \\ \hline
  0 & 4 & 3 & 2 & 2 & 1 & 1 & 1  \\
  1 & 0 & 0 & 0 & 1 & 1 & 2 & 1 \\
  \hline
\end{tabular}
\end{center}

By applying homology computation to each complex $\mathcal{R}_{\epsilon_i}$ in the filtration, PH tracks the evolution of topological features, such as connected components ($H_0$), loops ($H_1$), voids ($H_2$), and higher-dimensional analogues. A specific $p$-dimensional topological feature (a $p$-th homology class) is said to be "born" at scale $\epsilon_b$ if it first appears in $H_p(\mathcal{R}_{\epsilon_b})$. It subsequently "dies" at scale $\epsilon_d$ ($\epsilon_d > \epsilon_b$) when it merges with an older feature or becomes equivalent to zero (i.e., it gets "filled in" by higher-dimensional simplices). The interval $[\epsilon_b, \epsilon_d)$ represents the lifespan of this feature, and its length, $\epsilon_d - \epsilon_b$, is its persistence.

The collection of all such birth-death pairs $(\epsilon_b, \epsilon_d)$ for a given dimension $p$ is summarized in a persistence diagram (PD), a multiset of points in the extended plane $\overline{\mathbb{R}}^2$. Alternatively, these intervals can be visualized as a persistence barcode. Features with long persistence (points far from the diagonal $y=x$ in the PD) typically correspond to robust topological structures in the data, while short-lived features (points close to the diagonal) are often interpreted as noise or finer details. Formally, the inclusion map $\mathcal{R}_{\epsilon_i} \hookrightarrow \mathcal{R}_{\epsilon_j}$ for $\epsilon_i \le \epsilon_j$ induces a homomorphism between the $p$-dimensional homology groups, $f_{p}^{i,j}: H_p(\mathcal{R}_{\epsilon_i}) \to H_p(\mathcal{R}_{\epsilon_j})$. The $p$-dimensional persistent homology group $H_p^{i,j}$ is the image of this homomorphism, $H_p^{i,j} = \text{Im } f_{p}^{i,j}$, capturing the homology classes that persist from scale $\epsilon_i$ to $\epsilon_j$. This rigorous framework allows PH to provide a stable and informative summary of the underlying shape and structure of complex datasets across multiple scales.

\subsection{Comparing Topological Structures}

Beyond characterizing the topology of a single data manifold using persistent homology, several methods have been developed to compare the topological structures of two different datasets or representations, even when they reside in different ambient spaces. These methods often leverage the concept of barcodes but adapt them to capture relational topological information.

\subsubsection{Cross-Barcode for Comparing Data Manifolds}
To compare two data manifolds $M_{data}$ and $M_{model}$, represented by point clouds $P \subset \mathbb{R}^{D}$ and $Q \subset \mathbb{R}^{D}$ respectively, the Cross-Barcode$(P, Q)$ was introduced \cite{NEURIPS2021_3bc31a43}. This tool is designed to track multiscale topological discrepancies, essentially highlighting how the topology of $P$ differs relative to $Q$ across various distance scales \cite{NEURIPS2021_3bc31a43}.

The Cross-Barcode arises from the persistent homology of a specific filtered simplicial complex. This complex is the Vietoris-Rips complex built on the union of the point clouds, $P \cup Q$, but using a modified distance function. Specifically, let $m_{(P \cup Q)/Q}$ represent the pairwise distances where distances involving only points from $Q$ are set to zero, while other distances (within $P$, or between $P$ and $Q$) remain their standard Euclidean values . The filtration is built by considering increasing distance thresholds $\alpha$. The Cross-Barcode$(P, Q)$ then records the birth and death times of homology classes in the sequence of complexes $R_{\alpha}(\Gamma_{P\cup Q},m_{(P\cup Q)/Q})$ . Features that persist over a long range of $\alpha$ signify substantial topological differences between the manifolds represented by $P$ and $Q$ \cite{NEURIPS2021_3bc31a43}. The Manifold Topology Divergence (MTop-Div) score is often defined based on this, for example, as the sum of the lengths of all bars in the 1-dimensional Cross-Barcode \cite{NEURIPS2021_3bc31a43}.

\subsubsection{Representation Topology Divergence (RTD)}
When comparing two representations $P$ and $\tilde{P}$ (derived from the same underlying data $V$, so $|P| = |\tilde{P}| = N$) that might exist in different ambient spaces, the Representation Topology Divergence (RTD) provides a measure of topological dissimilarity, respecting the one-to-one correspondence between points \cite{pmlr-v162-barannikov22a}.

\begin{figure}[ht]
    \centering
    \includegraphics[width=0.8\linewidth]{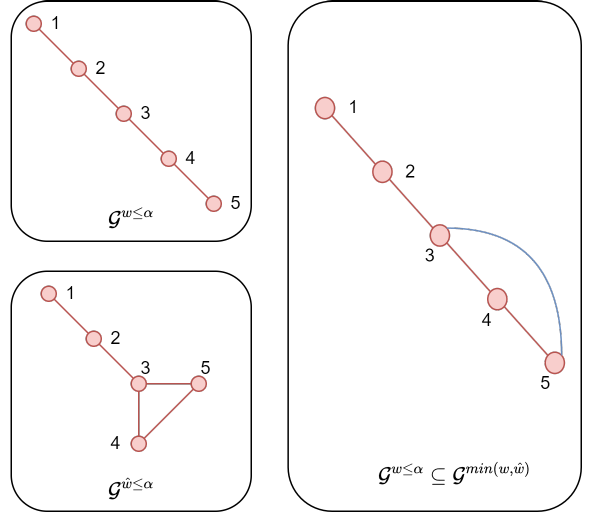}
    \caption{Graphs $\mathcal{G}^{w\leq\alpha}$, $\mathcal{G}^{\hat{w}\leq\alpha}$ and $\mathcal{G}^{min(w,\hat{w})\leq\alpha}$ with edges not in $\mathcal{G}^{w\leq\alpha}$ colored in blue}
    \label{fig:RTD}
\end{figure}

RTD is grounded in the R-Cross-Barcode$(P, \tilde{P})$. This barcode is computed from the persistent homology of an auxiliary weighted graph $\hat{\mathcal{G}}^{w,\tilde{w}}$. This graph has $2N$ vertices (conceptually, a copy $A_i$ for each original point in $P$ and another copy $A'_i$ for the corresponding point in $\tilde{P}$). The weights are defined to capture the comparison. A practical construction involves a $2N \times 2N$ matrix $m$ \cite{pmlr-v162-barannikov22a}:
\begin{equation}
m = \begin{pmatrix} 0 & (w_{+})^{T} \\ w_{+} & \min(w, \tilde{w}) \end{pmatrix}
\label{eq:rtd_matrix}
\end{equation}
where $w$ and $\tilde{w}$ are the $N \times N$ pairwise distance matrices within $P$ and $\tilde{P}$ respectively, $w_{+}$ is $w$ with its lower-triangular part set to $+\infty$ (effectively using only upper-triangular comparisons for one part), and $\min(w, \tilde{w})$ is the element-wise minimum of the two distance matrices . The R-Cross-Barcode is the persistence barcode of the Vietoris-Rips filtration built using this matrix $m$ .

The RTD score quantifies the magnitude of the topological discrepancies captured by the R-Cross-Barcode. It is typically defined as the sum of the lengths of all intervals $[b_i, d_i)$ in the 1-dimensional R-Cross-Barcode, often symmetrized \cite{pmlr-v162-barannikov22a}:

\begin{equation}
\begin{split} 
RTD(P, \tilde{P}) = \frac{1}{2} \Biggl( & \sum_{[b_i, d_i) \in \text{R-Cross-Barcode}_1(P, \tilde{P})} (d_i - b_i) \\
& + \sum_{[b'_j, d'_j) \in \text{R-Cross-Barcode}_1(\tilde{P}, P)} (d'_j - b'_j) \Biggr) 
\end{split}
\label{eq:rtd_definition}
\end{equation}
A key theoretical result is that if $RTD_k(P, \tilde{P}) = RTD_k(\tilde{P}, P) = 0$ for all dimensions $k \geq 1$, then $P$ and $\tilde{P}$ have identical standard persistence barcodes, dimension by dimension, and their topological features align spatially \cite{pmlr-v162-barannikov22a}. This makes RTD a strong candidate for enforcing topological similarity.

\subsubsection{Differentiable RTD for Topology-Preserving Representations}\label{RTDLoss}
To utilize RTD within gradient-based optimization for learning topology-preserving representations, its differentiability is crucial  \cite{trofimov2023learning} introduced a method to compute subgradients for the RTD score. The change in $RTD_k$ is linked to the filtration values $m_\sigma$ of the simplices $\sigma$ that cause the birth or death of homology classes \cite{trofimov2023learning}. The subgradient with respect to the filtration value of a simplex $\sigma$ can be expressed as:
\begin{equation}
\frac{\partial RTD_{k}(P, \tilde{P})}{\partial m_{\sigma}} = \sum_{i \in \mathcal{T}_k} \mathbb{I}\{f_k(d_i) = \sigma\} - \sum_{i \in \mathcal{T}_k} \mathbb{I}\{f_k(b_i) = \sigma\}
\label{eq:rtd_derivative_simplex}
\end{equation}
where $\mathcal{T}_k$ is the set of intervals $[b_i, d_i)$ in the R-Cross-Barcode$_k(P, \tilde{P})$, $f_k$ maps interval endpoints to the critical simplex responsible for the birth/death event, and $\mathbb{I}$ is the indicator function \cite{trofimov2023learning}. Using the chain rule, this allows computation of subgradients with respect to the coordinates of points in $P$ and $\tilde{P}$ \cite{trofimov2023learning}.

This differentiability enables the use of RTD as a regularization term in models like autoencoders (RTD-AE). The objective function combines a standard reconstruction loss with the RTD loss, weighted by a factor $\lambda$ \cite{trofimov2023learning}:
\begin{equation}
\mathcal{L}_{\text{RTD-AE}} = \frac{1}{2} \|X - X_{rec}\|^2 + \lambda RTD(X, Z)
\label{eq:rtdae_loss}
\end{equation}
Here, $X$ is the input data, $Z$ is the latent representation produced by the encoder, and $X_{rec}$ is the reconstructed data from the decoder \cite{trofimov2023learning}. Minimizing this loss encourages the network to learn representations $Z$ that are both faithful in reconstruction and topologically similar to the original high-dimensional data $X$ \cite{trofimov2023learning}.

\section{Related Work}
\subsection{Topological Data Analysis in Deep Learning}
Topological Data Analysis (TDA) \cite{edelsbrunner2010computational}, which employs algebraic tools like persistent homology to infer the topological structure of data spaces, has found numerous applications within deep learning. Its utility has been demonstrated across various domains, including graph machine learning \cite{rieck2019persistent,horn2021topological,zhao2020persistence,zhao2019learning,wong2021persistent,pham2025topological,southern2023curvature}, knowledge distillation \cite{kim2024do,jeon2024topological}, bioinformatics \cite{demir2022todd,luo2023improving,townsend2020representation}, image segmentation \cite{stucki2023topologically,hu2022learning,hu2021topology,hu2023learning,hu2019topology,clough2020topological,gupta2023topology}, and medical imaging analysis \cite{rieck2020uncovering,santhirasekaram2023topology,vandaele2023topological,nielson2015topological}, among others.

Several works have incorporated topological priors directly into the learning process, often through specialized loss functions. For instance, \cite{clough2020topological} utilized cubical homology to develop a topological loss function for image segmentation, enabling supervision based solely on topological priors without requiring ground-truth labels, applicable in semi-supervised or post-processing frameworks. Addressing medical imaging challenges, \cite{santhirasekaram2023topology} leveraged the characteristic limited structural variability between patients by combining a topological loss with vector quantization to enhance the robustness of segmentation models. \cite{gupta2023topology} introduced a topological loss aimed at enforcing identical Betti numbers between predicted segmentations and ground truth, providing a detailed explanation of the underlying Betti error mechanism.

TDA has also been employed to analyze the internal workings and representations of deep learning models. \cite{Rieck19a}, for example, used persistent homology to quantify the complexity of deep neural networks. \cite{pmlr-v162-barannikov22a} presented a TDA-based methodology for measuring dissimilarities between data representations, applicable both across different models and between layers within the same model. In the context of autoencoders, TDA helps ensure that latent representations retain essential topological features of the input data. \cite{pmlr-v119-moor20a} applied Vietoris-Rips complexes to compare the topology of the original data space with the latent space, encouraging the autoencoder to preserve multi-scale connectivity information. Along similar lines, \cite{trofimov2023learning} utilized Representative Topological Descriptors (RTD) to compare higher-dimensional topological features between the input and latent spaces of autoencoders.

TDA also contributes to knowledge transfer and model efficiency. \cite{kim2024do} demonstrated the use of persistence images to capture comprehensive geometric structures (e.g., distributional shape, multi-scale features, connectivity) for effective knowledge distillation from teacher to student models. Directly relevant to our work on Vision-Language Models (VLMs), \cite{zhang2024homology} proposed a Homology Consistency (HC) constraint to improve transfer learning efficiency. This method explicitly enforces structural equivalence, measured via persistent homology, between the latent manifolds of image and text representations during downstream fine-tuning.

\section{Methodology}

Our proposed method enhances few-shot transfer learning for Vision-Language Models (VLMs) like CLIP by integrating topological data analysis into the fine-tuning process. We hypothesize that while pre-trained VLMs possess substantial prior knowledge from large-scale datasets, a subtle distribution shift exists between this prior knowledge and the posterior distribution required for specific downstream\cite{radford_learning_2021,yu2023task}. Accurately estimating this shift using limited downstream samples is crucial for improving prediction accuracy. Beyond the direct semantic supervision provided by the standard Cross-Entropy (CE) loss, we posit that the topological structure of the shared latent space, jointly encoded by visual and textual representations through contrastive pre-training, contains valuable posterior information \cite{zhang2024homology}. Discrepancies in the topology between corresponding visual and textual representations can signal important task-specific variations. Our approach aims to leverage this topological information within an efficient fine-tuning framework.

\subsection{Task Residual Framework with Topological Regularization}

We adopt the Task Residual (TR) tuning framework as the foundation for our few-shot learning approach\cite{yu2023task}. The TR framework preserves the rich prior knowledge embedded in pre-trained VLMs by keeping the original encoder weights frozen during downstream tuning. Adaptation to the target task is achieved by learning additive residual parameters directly on the text-based classifier derived from the frozen text encoder.

To incorporate topological insights, we introduce a loss term based on Representation Topology Divergence (RTD)\cite{pmlr-v162-barannikov22a,trofimov2023learning}. RTD quantifies the dissimilarity in multi-scale topology between two point clouds of equal size that have a one-to-one correspondence, even if they reside in different ambient spaces. In our context, we apply RTD to measure the topological divergence between the batch-wise visual embeddings and their corresponding textual embeddings within the VLM's latent space. We integrate this topological loss ($L_{RTD}$, see section \ref{RTDLoss}) with the standard Cross-Entropy loss ($L_{CE}$) used in the TR framework. The total loss function for training is defined as:

\begin{equation}
L_{total} = L_{CE} + \lambda L_{RTD}
\label{eq:total_loss}
\end{equation}

where $\lambda$ is a hyperparameter weighting the contribution of the topological divergence term. $L_{CE}$ provides direct semantic supervision, while $L_{RTD}$ encourages the alignment of the topological structures of the visual and textual representations, capturing finer-grained posterior knowledge relevant to the downstream task.

\subsection{Training Strategy for Topological Alignment}

To effectively compute the RTD loss and leverage the TR framework, a specific training strategy is employed. Crucially, the pre-trained weights of both the visual and text encoders of the base VLM (e.g., CLIP) are kept frozen throughout the fine-tuning process. This ensures the preservation of the general prior knowledge learned during large-scale pre-training.

\begin{figure}[ht]
    \centering
    \includegraphics[width=0.9\linewidth]{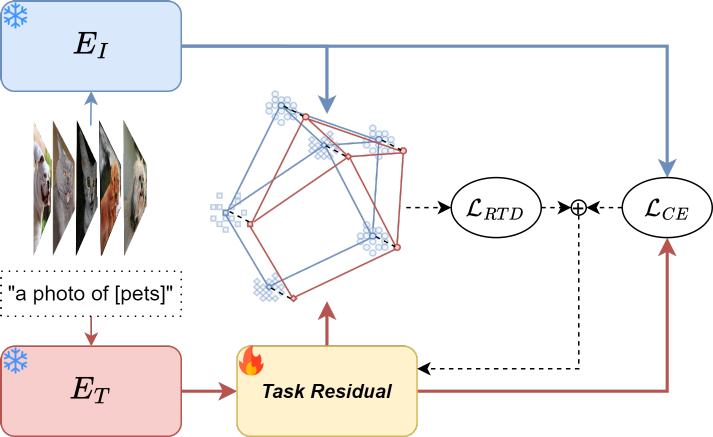}
    \caption{The proposed topologically-aware few-shot learning framework. The training loss combines Representation Topology Divergence ($L_{RTD}$) between visual and adapted text embeddings with Cross-Entropy ($L_{CE}$) to update only the Task Residual.}
    \label{fig:method}
\end{figure}

For efficient learning and to meet the requirements of RTD, we structure the training batches carefully. The batch size is set to be exactly equal to the number of classes ($K$) in the downstream dataset. Furthermore, each training batch is constructed such that it contains precisely one visual sample from each of the $K$ classes. This guarantees a one-to-one correspondence between the visual embeddings and the text embeddings (representing class prototypes or learned residuals) within each batch. This setup satisfies the prerequisites for calculating RTD, which requires paired point clouds\cite{pmlr-v162-barannikov22a}, and aligns with the efficient training methodology of the TR framework. This batch construction enables the effective comparison of topological structures between the visual and textual modalities for every training step, facilitating the optimization of the combined loss function (Eq. \ref{eq:total_loss}) and promoting the learning of topology-preserving representations beneficial for few-shot generalization.

\section{Experiments and Results}

\subsection{Experimental Settings}

Following standard practices in few-shot learning evaluation for VLMs, we assess our method on 6 diverse benchmark datasets: OxfordPets\cite{parkhi2012cats}, Food101\cite{bossard2014food}, FGVCAircraft\cite{maji2013fine}, EuroSAT\cite{helber2019eurosat}, Caltech101\cite{fei2004learning}, and DTD\cite{cimpoi2014describing}. These datasets encompass a variety of visual classification challenges. For training, we randomly sample \{1, 2, 4, 8, 16\} images per class from the respective training sets and evaluate performance on the full test sets.
\begin{figure*}[htbp] 
    \centering 
    \includegraphics[width=\textwidth]{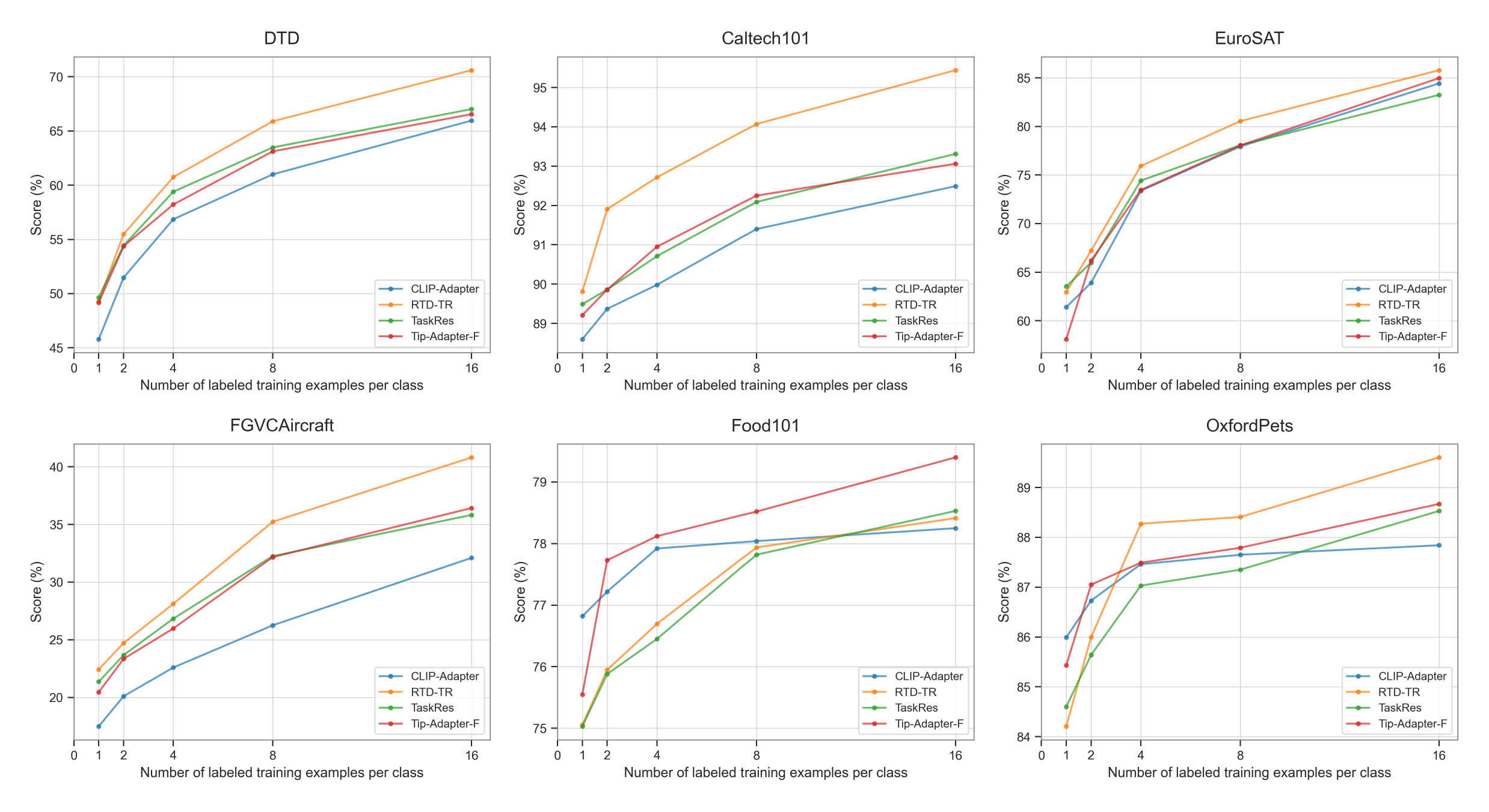} 
    \caption{The performance comparison of our proposed topologically-aware method (RTD-TR) against baseline approaches (CLIP-Adapter, TaskRes, Tip-Adapter-F) on few-shot learning tasks. Results include accuracy for 1, 2, 4, 8, and 16 shots per class across 6 benchmark datasets.} 
    \label{fig:result} 
\end{figure*}

Consistent with our methodology, the batch size during training is set equal to the number of classes ($K$) for each specific dataset. Every batch is constructed to contain exactly one sample from each of the $K$ classes, ensuring the one-to-one correspondence necessary for the RTD loss calculation and aligning with the Task Residual framework.

We employ the Adam optimizer with an initial learning rate of $1 \times 10^{-4}$. The learning rate is decayed using a cosine annealing schedule. The hyperparameter $\lambda$, which balances the Cross-Entropy loss ($\mathcal{L}_{CE}$) and the RTD loss ($\mathcal{L}_{RTD}$), is critical. Preliminary experiments indicated optimal performance when the initial loss ratio $\lambda \mathcal{L}_{RTD} / \mathcal{L}_{CE}$ falls within the range [0.33, 0.37]. Therefore, we utilize a binary search strategy to determine the most effective $\lambda$ value for each specific dataset and shot combination. We compare our method, denoted RTD-TR, against relevant baselines including CLIP-Adapter, the base TaskRes framework, and Tip-Adapter-F.

\subsection{Performance Analysis}
As illustrated in Figure \ref{fig:result}, our proposed method, RTD-TR, consistently demonstrates strong performance across the six benchmark datasets and various few-shot settings (1, 2, 4, 8, and 16 shots).

Compared to its base framework, TaskRes, RTD-TR achieves noticeable improvements on nearly all datasets and shot counts. This highlights the effectiveness of incorporating the Representation Topology Divergence loss, which successfully leverages the topological information in the latent space to enhance few-shot adaptation. Similarly, RTD-TR consistently outperforms the CLIP-Adapter baseline across all evaluated scenarios.

When compared with Tip-Adapter-F, which is a strong few-shot learning baseline, RTD-TR shows competitive results. While Tip-Adapter-F exhibits leading performance on several datasets, particularly fine-grained ones like FGVCAircraft, Food101, and OxfordPets, RTD-TR often closes the gap or surpasses it as the number of shots increases. For instance, on DTD and Caltech101, RTD-TR is highly competitive and achieves the best performance among the compared methods at 16 shots. On EuroSAT, RTD-TR performs comparably to Tip-Adapter-F.

Overall, the results validate our hypothesis that aligning the topological structures of visual and textual representations via RTD within an efficient transfer learning framework like Task Residual leads to enhanced performance in few-shot image classification. The consistent gains over the TaskRes baseline underscore the benefit derived specifically from the topological constraint.

\section{Conclusions}
\label{sec:conclusions}

In this paper, we addressed the challenge of efficiently adapting large-scale Vision-Language Models (VLMs) to downstream classification tasks under few-shot constraints. We argued that merely relying on semantic similarity objectives like Cross-Entropy might not fully capture the necessary task-specific information, particularly when dealing with the subtle distribution shifts encountered during transfer learning. We proposed leveraging the topological structure of the VLM's shared latent space as an additional source of information.

Our core contribution is a topologically-aware tuning framework built upon Task Residual (TR) tuning. We introduced Representation Topology Divergence (RTD) as a loss term to explicitly measure and minimize the topological discrepancies between visual embeddings and their corresponding adapted textual representations within each training batch. By freezing the powerful pre-trained encoders and optimizing only the additive Task Residual parameters using a combined objective ($\mathcal{L}_{CE} + \lambda \mathcal{L}_{RTD}$), our method effectively preserves prior knowledge while promoting topological alignment relevant to the downstream task. A specific batching strategy (batch size = number of classes, one sample per class) was employed to enable the required one-to-one correspondence for RTD calculation.

Experimental results on six diverse benchmark datasets confirmed the efficacy of our approach. Our method, RTD-TR, consistently outperformed the standard TaskRes baseline and CLIP-Adapter, demonstrating the tangible benefits of incorporating topological regularization. Furthermore, it showed competitive performance against strong baselines like Tip-Adapter-F, particularly in scenarios with more available shots or on datasets where topological structure might play a more significant role.

This work underscores the potential of using topological data analysis not just for understanding representations, but also for actively improving model adaptation. Aligning the multi-scale topological features of visual and text modalities provides a complementary signal to traditional semantic losses, potentially leading to more robust and generalizable few-shot learners. While the computational cost of RTD warrants consideration, the demonstrated performance gains suggest this is a promising direction. Future work could explore more computationally efficient topological metrics, investigate the role of higher-dimensional homology, extend the approach to other VLM architectures and downstream tasks beyond classification, and further analyze the theoretical connections between topological similarity and few-shot generalization in VLMs.




\bibliography{paper}
\bibliographystyle{aaai}

\end{document}